%% file: multi30k.tex
\documentclass[11pt]{article}
\usepackage{acl2016}
\usepackage{times}
\usepackage{latexsym}
\usepackage[usenames,dvipsnames]{xcolor}
\usepackage[utf8]{inputenc}
\usepackage{url}
\usepackage{amsmath}
\usepackage{subcaption}
\usepackage{booktabs}
\usepackage{graphicx}
\usepackage[official]{eurosym}
\usepackage[compact]{titlesec}
\newcommand{\dataset}{\textbf{Multi30K}}

\title{\dataset{}: Multilingual English-German Image Descriptions}

\aclfinalcopy 

\author{
  Desmond Elliott \and Stella Frank \and Khalil Sima'an\\
  ILLC, University of Amsterdam\\
  {\tt \{d.elliott, s.c.frank, k.simaan\}}{\tt @uva.nl}
  \AND
  Lucia Specia\\
  Sheffield University\\
  {\tt l.specia@sheffield.ac.uk}
  }

\date{}

\begin{document}

\maketitle

\input{abstract}
\input{intro}
\input{dataset}
\input{discussion}
\input{conclusions}

\section*{Acknowledgements}

DE and KS were supported by the NWO Vici grant nr.\
277-89-002. SF was supported by European Union’s Horizon 2020 research
and innovation programme under grant agreement nr.\ 645452. We are
grateful to Philip Schulz for checking the German translation of the
worker instructions, and to Joachim Daiber for providing the
pre-trained truecasing models.

\bibliography{library}
\bibliographystyle{acl2016}

\end{document}

%% file: abstract.tex
\begin{abstract}

  We introduce the \dataset{} dataset to stimulate multilingual
  multimodal research.
  Recent advances in image description have been demonstrated on
  English-language datasets almost exclusively, but image
  description should not be limited to English.
  This dataset extends the Flickr30K dataset with i) German
  translations created by professional translators over a subset of
  the English descriptions, and ii) descriptions crowdsourced
  independently of the original English descriptions. We outline how
  the data can be used for multilingual image description and
  multimodal machine translation, but we anticipate the data will be
  useful for a broader range of tasks.

\end{abstract}

%% file: intro.tex
\section{Introduction}

Image description is one of the core challenges at the intersection of
Natural Language Processing (NLP) and Computer Vision (CV) \cite{Bernardi2016}.
This task has only received attention in a monolingual English
setting, helped by the availability of English datasets, e.g.
Flickr8K \cite{Hodosh2013}, Flickr30K \cite{Young2014}, and MS COCO
\cite{Chen2015}. However, the possible applications of image
description are useful for all languages, such as searching for images
using natural language, or providing alternative-description text for
visually impaired Web users.

We introduce a large-scale dataset of images paired with sentences in
English and German as an initial step towards studying the value and
the characteristics of multilingual-multimodal data. \dataset{} is an
extension of the Flickr30K dataset \cite{Young2014} with 31,014 German
\textit{translations} of English descriptions and 155,070
independently collected German descriptions.  The translations were
collected from professionally contracted translators, whereas the
descriptions were collected from untrained crowdworkers. The key
difference between these corpora is the relationship between the
sentences in different languages.  In the translated corpus, we know
there is a strong correspondence between the sentences in both
languages.  In the descriptions corpus, we only know that the
sentences, regardless of the language, are supposed to describe the
same image.

A dataset of images paired with sentences in multiple languages
broadens the scope of multimodal NLP research. Image description with
\textit{multilingual} data can also be seen as machine translation in
a \textit{multimodal} context.  This opens up new avenues for
researchers in machine translation
\cite[\textit{inter-alia}]{Koehn2003,Chiang2005,Sutskever2014,Bahdanau2015}
to work with multilingual multimodal data.  Image--sentence ranking
using monolingual multimodal datasets
\cite[\textit{inter-alia}]{Hodosh2013} is also a natural task for
multilingual modelling. 

The only existing datasets of images
paired with multilingual sentences were created by
professionally translating English into the target language: IAPR-TC12 with 20,000
English-German described images \cite{Grubinger2006}, and the Pascal
Sentences Dataset of 1,000 Japanese-English described images
\cite{Funaki2015}. \dataset{} dataset is larger than both of these and contains
both independent and translated sentences. We hope this dataset will
be of broad interest across NLP and CV research and anticipate that
these communities will put the data to use in a broader range of tasks
than we can foresee.

%% file: dataset.tex
\input{examples-figure}
\section{The \dataset{} Dataset}\label{sec:data}

The Flickr30K Dataset contains 31,014 images sourced from online
photo-sharing websites \cite{Young2014}. Each image is paired with
five English descriptions, which were collected from Amazon Mechanical
Turk\footnote{\url{http://www.mturk.com}}. The dataset contains 145,000
training, 5,070 development, and 5,000 test descriptions.  The \dataset{}
dataset extends the Flickr30K dataset with \textit{translated} and
\textit{independent} German sentences.

\subsection{Translations}\label{sec:data:translations}

The translations were collected from professional
English-German translators contracted via an established Language
Service in Germany. Figure
\ref{fig:dataset:example} presents an example of the differences
between the types of data.
We collected one translated description per image,
resulting in a total of 31,014 translations.
To ensure an even distribution over description length, the English
descriptions were chosen based on their relative length, with an equal
number of longest, shortest, and median length source descriptions.
We paid a total of \euro 23,000 to collect the data (\euro 0.06 per
word). Translators were shown an English language
sentences and asked to produce a correct and fluent translation for it
in German, without seeing the image.
We decided against showing the images to translators to
make this as close as possible to a standard translation task,
also making the data collected here distinct from the
independent descriptions collected as described in Section
\ref{sec:data:descriptions}.

\subsection{Independent Descriptions}\label{sec:data:descriptions}

\begin{figure*}[t]
    \centering
    \includegraphics[scale=0.15]{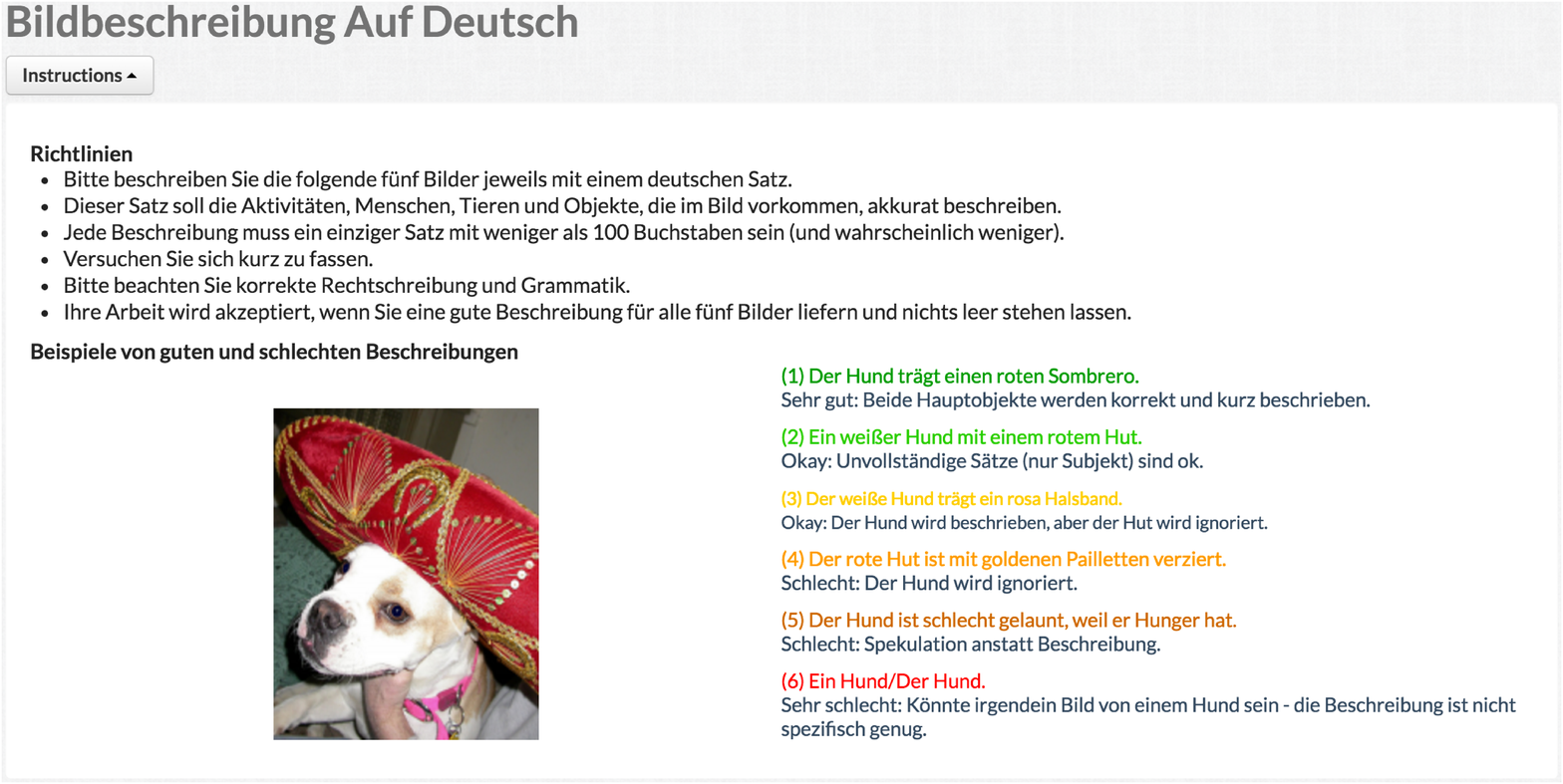}

    \caption{The German instructions shown to crowdworkers were
    translated from the original instructions.}\label{fig:dataset:crowdflower}

\end{figure*}

The descriptions were collected from crowdworkers via the
Crowdflower platform\footnote{\url{http://www.crowdflower.com}
}.  We collected five descriptions per image in the Flickr30K dataset,
resulting in a total of 155,070 sentences. Workers were presented with
a translated version of the data collection interface used by
\cite{Hodosh2013}, as shown in Figure \ref{fig:dataset:crowdflower}.
We translated the interface to make the task as similar as
possible to the crowdsourcing of the English sentences. The
instructions were translated by one of the authors and checked by a
native German Ph.D student.

185 crowdworkers took part in the task over a period of 31 days. We
split the task into 1,000 randomly selected images per day to control
the quality of the data and to prevent worker fatigue.  Workers were
required to have a German-language skill certification and be at least
a Crowdflower Level 2 Worker: they have participated in at least 10
different Crowdflower jobs, has passed at least 100 quality-control
questions, and has an job acceptance rate of at least 85\%.

The descriptions were collected in batches of five images per job.
Each image was randomly selected from the complete set of 1,000 images
for that day, and workers were limited to writing at most 250
descriptions per day. We paid workers \$0.05 per
description\footnote{This is the same rate as \newcite{Rashtchian2010}
and \newcite{Elliott2013} paid to collect English sentences.} and
limited them submitting faster than 90s per job to discourage
poor/low-quality work.  (This works out at a rate of 40 jobs per hour,
i.e. 200 descriptions / hour.) We configured Crowdflower to
automatically ban users who worked faster than this rate. Thus the
theoretical maximum wage per hour was \$10/hour. We paid a total of
\$9,591.24 towards collecting the data and paying the Crowdflower
platform fees.

During the collection of the data, we assessed the quality both by
manually checking a subset of the descriptions and also with automated
checks. We inspected the submissions of users who wrote sentences with
less than five words, and users with high type to token ratios (to
detect repetition). We also used a character-level 6-gram LM to flag
descriptions with high perplexity, which was very effective at
catching nonsense sentences. In general we did not have to ban or
reject many users and overall description quality was high.

\subsection{Translated vs. Independent Descriptions}\label{sec:data:t_v_i}

We now analyse the differences between the translated and the
description corpora.  For this analysis, all sentences were stripped
of punctuation and truecased using the Moses \texttt{truecaser.pl}
script trained over Europarl v7 and News Commentary v11 English-German
parallel corpora.

Table \ref{tab:dataset:comparison} shows the differences between the
corpora. The German translations are longer than the independent
descriptions (11.1 vs.\ 9.6 words), while the English descriptions
selected for translation are slightly shorter, on average, than the
Flickr30k average (11.9 vs.\ 12.3).
When we compare the German
translation dataset against an equal number of sentences from the
German descriptions dataset,
we find that the translations also have more word types (19.3K vs.\
17.6K), and more singleton types occurring only once (11.3K vs.\
10.2K; in both datasets singletons comprise 58\% of the vocabulary).
The translations thus have a wider vocabulary, despite being generated
by a smaller number of authors.
The English datasets (all descriptions vs. those selected for
translation) show a similar trend, indicating that these differences
may be a result of
the decision to select equal numbers of short,
medium, and long English sentences for translation.

\input{comparison-table}

\subsection{English vs. German}\label{sec:data:e_v_d}

The English image descriptions are generally longer than the German
descriptions, both in terms of number of words and characters. Note
that the difference is much less smaller when measuring characters:
German uses 22\% fewer words but only 2.5\% fewer characters. However,
we observe a different pattern in the translation corpora: German uses
6.6\% fewer words than English but 17.1\% more characters.  The
vocabulary of the German description and translation corpora are more
than twice as large as the English corpora.  Additionally, the German
corpora have two-to-three times as many singletons. This is likely due
to richer morphological variation in German, as well as word
compounding.

%% file: examples-figure.tex
\begin{figure*}
  \begin{subfigure}[t!]{0.4\textwidth}
  \begin{enumerate}
    \item Brick layers constructing a wall.\\
    \item Maurer bauen eine Wand.\\
  \end{enumerate}
  \end{subfigure}%
  \hspace{1em}%
  \begin{subfigure}[c]{0.15\textwidth}
  \centering
  \includegraphics[scale=0.125]{figures/2158925258.eps}
  \end{subfigure}%
  \hspace{1em}%
  \begin{subfigure}[c]{0.4\textwidth}
  \begin{enumerate}
    \item The two men on the scaffolding are helping to build a red brick wall.
    \item Zwei Mauerer mauern ein Haus zusammen.
  \end{enumerate}
  \end{subfigure}

  \vspace{2em}

  \begin{subfigure}[t!]{0.4\textwidth}
  \begin{enumerate}
    \item Trendy girl talking on her cellphone while gliding slowly down the street
    \item Ein schickes Mädchen spricht mit dem Handy
      während sie langsam die Straße entlangschwebt.
  \end{enumerate}
  \subcaption{Translations}
  \end{subfigure}%
  \hspace{1em}%
  \begin{subfigure}[c]{0.15\textwidth}
  \centering
  \includegraphics[scale=0.19]{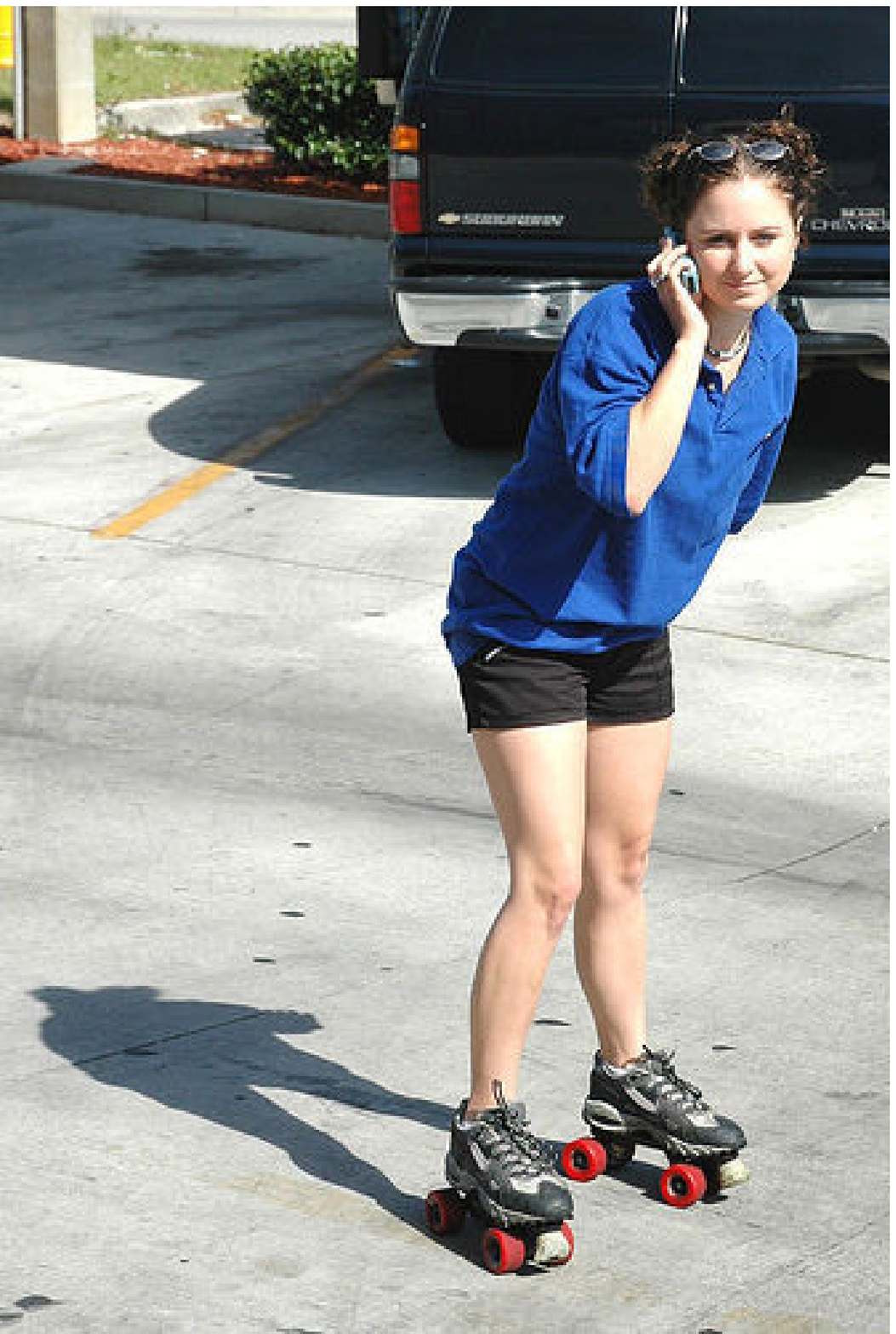}
  \end{subfigure}%
  \hspace{1em}%
  \begin{subfigure}[c]{0.4\textwidth}
  \begin{enumerate}
    \item There is a young girl on her cellphone while skating.
    \item Eine Frau im blauen Shirt telefoniert beim Rollschuhfahren.\\
  \end{enumerate}
  \caption{Independent descriptions}
  \end{subfigure}

  \caption{Multilingual examples in the \dataset{} dataset.
  The independent
  sentences are all accurate descriptions of the image but do not
  contain the same details in both languages, such as shirt colour or
  the scaffolding.
  In the second translation pair (bottom left) the translator has
  translated ``glide'' as ``schweben'' (``to float'') probably due to
  not seeing the image context (see Section
  \ref{sec:data:translations} for more details). 
  }\label{fig:dataset:example}

\end{figure*}

%% file: comparison-table.tex
%

%

\begin{table*}
\renewcommand{\arraystretch}{1.1}
\centering
\begin{tabular}{llrrrrr}
\toprule
                						& & Tokens 		& Types 	& Characters 		& Avg. length & Singletons\\
\midrule
\multicolumn{2}{l}{Translations (31,014 sentences)} & & & &  & \\
&English 		& 357,172 	& 11,420  	& 1,472,251 	& 11.9            & 5,073\\
&German      & 333,833	& 19,397 	& 1,774,234 	& 11.1            & 11,285    \\
\midrule
\multicolumn{2}{l}{Descriptions (155,070 sentences)} & & & & & \\
&English & 1,841,159 	& 22,815 	& 7,611,033	& 12.3            & 9,230 \\
&German & 1,434,998 	& 46,138 	& 7,418,572 	& 9.6              & 26,510  \\
\bottomrule
\end{tabular}
\caption{Corpus-level statistics about the translation and the
description data.}
\label{tab:dataset:comparison}
\end{table*}

%% file: discussion.tex
\section{Discussion}\label{sec:discusison}

The \dataset{} dataset is immediately suitable for research on a wide
range of tasks, including but not limited to automatic image
description, image--sentence ranking, multimodal and multilingual
semantics, and machine translation.

\subsection{\dataset{} for Image Description}

Deep neural networks for image description typically integrate visual
features into a recurrent neural network language model
\cite[\textit{inter-alia}]{Vinyals2015,Xu2015}.
\newcite{ElliottFrankHasler2015} demonstrated how to build
multilingual image description models that learn and transfer features
between monolingual image description models. They performed a series
of experiments on the IAPR-TC12 dataset \cite{Grubinger2006} of images
aligned with German translations, showing that both English and German
image description could be improved by transferring features from a
multimodal neural language model trained to generate descriptions in
the other language. The \dataset{} dataset will enable further
research in this direction, allowing researchers to work with larger
datasets with multiple references per image.

\subsection{\dataset{} for Machine Translation}

Machine translation is typically performed using only textual data,
for example news data, the Europarl corpora, or corpora harvested from
the Web (CommonCrawl, Wikipedia, etc.). The \dataset{} dataset makes
it possible to further develop machine translation in a setting where
multimodal data, such as images or video, are observed alongside text.
The potential advantages of using multimodal information for machine
translation include the ability to better deal with ambiguous source
text and to avoid (untranslated) out-of-vocabulary words in the target
language \cite{calixto-etal_VLNET:2012}.  \newcite{Hitschler2016} have
demonstrated the potential of multimodal features in a target-side
translation reranking model.  Their approach is initially trained over
large text-only translation copora and then fine-tuned with a small
amount of in-domain data, such as our dataset. We expect a
variety of translation models can be adapted to take advantage
of multimodal data as features in a log-linear model or as feature
vectors in neural machine translation models.

%% file: conclusions.tex
\section{Conclusions}\label{sec:conclusions}

We introduced \dataset{}: a large-scale multilingual multimodal
dataset for interdisciplinary machine learning research. Our dataset
is an extension of the popular Flickr30K dataset with descriptions and
professional translations in German. 

The descriptions were collected from a crowdsourcing platform, while
the translations were collected from professionally contracted
translators. These differences are deliberate and part of the larger
scope of studying multilingual multimodal data in different contexts.
The descriptions were collected as similarly as possible to the
original Flickr30K dataset by translating the instructions used by
\newcite{Young2014} into German. The translations were collected
without showing the images to the translators to keep it as close to a
standard translation task as possible.

There are substantial differences between the translated and the
description datasets. The translations contain approximately
the same number of tokens and have sentences of approximately the same
length in both languages. These properties make them suited to machine
translations models. The description datasets are
very different in terms of average sentence lengths and the number
of word types per language. This is likely to cause different
engineering and scientific challenges because the
descriptions are independently collected corpora instead of a
sentence-level aligned corpus.

In the future, we want to study multilingual multimodality over a
wider range of languages, for example beyond Indo-European families.
We call on the community to engage with us on creating massively
multilingual multimodal datasets.

%% file: multi30k.bbl
\begin{thebibliography}{}

\bibitem[\protect\citename{Bahdanau \bgroup et al.\egroup }2015]{Bahdanau2015}
Dzmitry Bahdanau, Kyunghyun Cho, and Yoshua Bengio.
\newblock 2015.
\newblock Neural machine translation by jointly learning to align and
  translate.
\newblock In {\em Proceedings of ICLR}.

\bibitem[\protect\citename{Bernardi \bgroup et al.\egroup }2016]{Bernardi2016}
Raffaella Bernardi, Ruken Cakici, Desmond Elliott, Aykut Erdem, Erkut Erdem,
  Nazli Ikizler-Cinbis, Frank Keller, Adrian Muscat, and Barbara Plank.
\newblock 2016.
\newblock Automatic description generation from images: A survey of models,
  datasets, and evaluation measures.
\newblock {\em Journal of Artificial Intelligence Research}, 55:409--442.

\bibitem[\protect\citename{Calixto \bgroup et al.\egroup
  }2012]{calixto-etal_VLNET:2012}
Iacer Calixto, Teo~E. de~Campos, and Lucia Specia.
\newblock 2012.
\newblock Images as context in statistical machine translation.
\newblock In {\em Second Annual Meeting of the EPSRC Network on Vision \&
  Language}, Sheffield.

\bibitem[\protect\citename{Chen \bgroup et al.\egroup }2015]{Chen2015}
Xinlei Chen, Hao Fang, Tsung{-}Yi Lin, Ramakrishna Vedantam, Saurabh Gupta,
  Piotr Doll{\'{a}}r, and C.~Lawrence Zitnick.
\newblock 2015.
\newblock Microsoft {COCO} captions: Data collection and evaluation server.
\newblock {\em CoRR}, abs/1504.00325.

\bibitem[\protect\citename{Chiang}2005]{Chiang2005}
David Chiang.
\newblock 2005.
\newblock A hierarchical phrase-based model for statistical machine
  translation.
\newblock In {\em Proceedings of the 43rd Annual Meeting on Association for
  Computational Linguistics}, pages 263--270. Association for Computational
  Linguistics.

\bibitem[\protect\citename{Elliott and Keller}2013]{Elliott2013}
Desmond Elliott and Frank Keller.
\newblock 2013.
\newblock {Image Description using Visual Dependency Representations}.
\newblock In {\em EMNLP}, pages 1292--1302.

\bibitem[\protect\citename{Elliott \bgroup et al.\egroup
  }2015]{ElliottFrankHasler2015}
Desmond Elliott, Stella Frank, and Eva Hasler.
\newblock 2015.
\newblock Multi-language image description with neural sequence models.
\newblock {\em CoRR}, abs/1510.04709.

\bibitem[\protect\citename{Funaki and Nakayama}2015]{Funaki2015}
Ruka Funaki and Hideki Nakayama.
\newblock 2015.
\newblock Image-mediated learning for zero-shot cross-lingual document
  retrieval.
\newblock In {\em Empirical Methods in Natural Language Processing 2015}, pages
  585--590.

\bibitem[\protect\citename{Grubinger \bgroup et al.\egroup
  }2006]{Grubinger2006}
Michael Grubinger, Paul~D. Clough, Henning Muller, and Thomas Desealers.
\newblock 2006.
\newblock The {IAPR TC-12} benchmark: A new evaluation resource for visual
  information systems.
\newblock In {\em LREC}.

\bibitem[\protect\citename{Hitschler and Riezler}2016]{Hitschler2016}
Julian Hitschler and Stefan Riezler.
\newblock 2016.
\newblock Multimodal pivots for image caption translation.
\newblock {\em CoRR}, abs/1601.03916.

\bibitem[\protect\citename{Hodosh \bgroup et al.\egroup }2013]{Hodosh2013}
Micah Hodosh, Peter Young, and Julia Hockenmaier.
\newblock 2013.
\newblock {Framing Image Description as a Ranking Task: Data, Models and
  Evaluation Metrics}.
\newblock {\em Journal of Artificial Intelligence Research}, 47:853--899.

\bibitem[\protect\citename{Koehn \bgroup et al.\egroup }2003]{Koehn2003}
Philipp Koehn, Franz~Josef Och, and Daniel Marcu.
\newblock 2003.
\newblock Statistical phrase-based translation.
\newblock In {\em Proceedings of the 2003 Conference of the North American
  Chapter of the Association for Computational Linguistics on Human Language
  Technology - Volume 1}, NAACL '03, pages 48--54, Stroudsburg, PA, USA.
  Association for Computational Linguistics.

\bibitem[\protect\citename{Rashtchian \bgroup et al.\egroup
  }2010]{Rashtchian2010}
Cyrus Rashtchian, Peter Young, Micha Hodosh, and Julia Hockenmaier.
\newblock 2010.
\newblock Collecting image annotations using {A}mazon's {M}echanical {T}urk.
\newblock In {\em NAACLHLT Workshop on Creating Speech and Language Data with
  {A}mazon's {M}echanical {T}urk}.

\bibitem[\protect\citename{Sutskever \bgroup et al.\egroup
  }2014]{Sutskever2014}
Ilya Sutskever, Oriol Vinyals, and Quoc V.~V Le.
\newblock 2014.
\newblock Sequence to sequence learning with neural networks.
\newblock In Z.~Ghahramani, M.~Welling, C.~Cortes, N.D. Lawrence, and K.Q.
  Weinberger, editors, {\em Advances in Neural Information Processing Systems
  27}, pages 3104--3112. Curran Associates, Inc.

\bibitem[\protect\citename{Vinyals \bgroup et al.\egroup }2015]{Vinyals2015}
Oriol Vinyals, Alexander Toshev, Samy Bengio, and Dumitru Erhan.
\newblock 2015.
\newblock Show and tell: A neural image caption generator.
\newblock In {\em The IEEE Conference on Computer Vision and Pattern
  Recognition (CVPR)}, June.

\bibitem[\protect\citename{Xu \bgroup et al.\egroup }2015]{Xu2015}
Kelvin Xu, Jimmy Ba, Ryan Kiros, Kyunghyun Cho, Aaron~C. Courville, Ruslan
  Salakhutdinov, Richard~S. Zemel, and Yoshua Bengio.
\newblock 2015.
\newblock Show, attend and tell: Neural image caption generation with visual
  attention.
\newblock {\em CoRR}, abs/1502.03044.

\bibitem[\protect\citename{Young \bgroup et al.\egroup }2014]{Young2014}
Peter Young, Alice Lai, Micha Hodosh, and Julia Hockenmaier.
\newblock 2014.
\newblock From image descriptions to visual denotations: New similarity metrics
  for semantic inference over event descriptions.
\newblock {\em Transactions of the Association for Computational Linguistics}.

\end{thebibliography}
